\documentclass[letterpaper, 10pt, conference]{template/ieeeconf}  

\IEEEoverridecommandlockouts                              

\usepackage[utf8]{inputenc} 
\usepackage[T1]{fontenc}    
\usepackage{hyperref}       
\usepackage{url}            
\usepackage{booktabs}       
\usepackage{amsfonts}       
\usepackage{nicefrac}       
\usepackage{microtype}      
\usepackage{xcolor}         

\usepackage{amsmath}
\usepackage{makecell}
\usepackage{graphicx}
\usepackage{multirow}

\usepackage{svg}

\usepackage{cite}
\usepackage{acro}
\acsetup{patch/maketitle=false}
\DeclareAcronym{RL}{
    short = RL,
    long = reinforcement learning
}

\DeclareAcronym{IL}{
    short = IL,
    long = imitation learning
}

\DeclareAcronym{IDM}{
    short = IDM,
    long =  intelligent driver model
}

\DeclareAcronym{OVM}{
    short = OVM,
    long =  optimal velocity model
}

\DeclareAcronym{MOBIL}{
    short = MOBIL,
    long =  minimizing overall braking induced by lane changes,
}

\DeclareAcronym{AV}{
    short = AV,
    long =  autonomous vehicle,
}

\DeclareAcronym{ALKS}{
    short = ALKS,
    long =  active lane keeping system,
}

\DeclareAcronym{LLM}{
    short = LLM,
    long =  large language model,
}

\DeclareAcronym{GAIL}{
    short = GAIL,
    long =  generative adversarial imitation learning,
}

\DeclareAcronym{BC}{
    short = BC,
    long =  behavior cloning,
}

\DeclareAcronym{SOTIF}{
    short = SOTIF,
    long =  safety of the intended functionality
}

\DeclareAcronym{VAE}{
    short = VAE,
    long =  variational autoencoder
}

\DeclareAcronym{DGM}{
    short = DGM,
    long =  deep generative model
}

\DeclareAcronym{DAR}{
    short = DAR,
    long =  deep autoregressive model
}

\DeclareAcronym{GNN}{
    short = GNN,
    long =  graph neural network
}

\DeclareAcronym{ADE}{
    short = ADE,
    long =  average distance error 
}

\DeclareAcronym{FDE}{
    short = FDE,
    long =  final distance error 
}

\DeclareAcronym{TTC}{
    short = TTC,
    long =  time to collision 
}

\DeclareAcronym{RSS}{
    short = RSS,
    long =  responsibility-sensitive safety
}

\DeclareAcronym{MAAE}{
    short = MAAE,
    long =  mean absolute acceleration error
}

\DeclareAcronym{MASE}{
    short = MASE,
    long =  mean absolute steering error
}

\begin{document}

\bstctlcite{IEEEexample:BSTcontrol}

\title{\LARGE \bf
Adversarial and Reactive Traffic Entities for Behavior-Realistic \\ Driving Simulation: A Review
}

\author{Joshua Ransiek$^{1}$, Philipp Reis$^{1}$, Tobias Schürmann$^{1}$, and Eric Sax$^{1}$
\thanks{$^{1}$are with the FZI Research Center for Information Technology, Karlsruhe, Germany
{\tt\small ransiek@fzi.de}}%
}

\maketitle
\thispagestyle{empty}
\pagestyle{empty}

\begin{abstract}
Despite advancements in perception and planning for \acp{AV}, validating their performance remains a significant challenge.
The deployment of planning algorithms in real-world environments is often ineffective due to discrepancies between simulations and real traffic conditions. 
Evaluating \acp{AV} planning algorithms in simulation typically involves replaying driving logs from recorded real-world traffic.
However, entities replayed from offline data are not reactive, lack the ability to respond to arbitrary \ac{AV} behavior, and cannot behave in an adversarial manner to test certain properties of the driving policy. 
Therefore, simulation with realistic and potentially adversarial entities represents a critical task for \ac{AV} planning software validation. 
In this work, we aim to review current research efforts in the field of traffic simulation, focusing on the application of advanced techniques for modeling realistic and adversarial behaviors of traffic entities.
The objective of this work is to categorize existing approaches based on the proposed classes of traffic entity behavior and scenario behavior control.  
Moreover, we collect traffic datasets and examine existing traffic simulations with respect to their employed default traffic entities.
Finally, we identify challenges and open questions that hold potential for future research.
\end{abstract}

\section{Introduction}

Simulation is an essential tool for safe, low-cost development and testing of \acf{AV} software. 
Core aspects of an effective simulation include photorealistic rendering \cite{cai_summit_2020, osinski_carla_2020, amini_vista_2022} and behaviorally realistic dynamic entities \cite{gulino_waymax_2024, li_metadrive_2022}. 
Behavior-realistic refers to the incorporation of behaviors for simulated entities other than the \ac{AV} that closely emulate the traffic the \ac{AV} encounters in real-world scenarios.    
A common approach to achieving this is by replaying pre-recorded human-driven trajectories, where the movements of other traffic participants are simulated around the \ac{AV} exactly as they occurred during data collection. 
However, simply replaying pre-recorded data can lead to pose divergence or simulation drift. Simulation drift \cite{bergamini_simnet_2021}, \cite{montali_waymo_2024} refers to the deviation between the \ac{AV}'s behavior in the pre-recorded data and its behavior during simulation. 
If the \ac{AV}’s actions diverge from those in the recorded data, other traffic participants fail to react accordingly, making the simulation unrealistic and ineffective for validation. 
Therefore, a key requirement for modeling realistic behavior in simulation is the ability for simulated traffic agents to respond to the arbitrary behavior of the \ac{AV}.
Rule-based methods \cite{treiber_congested_2000,kreutz_analysis_2021,sugiyama_optimal_1999,kesting_general_2007} equip entities with a rule-based behavior function in order to react to surrounding traffic participants. 
To further enhance the realism of simulated entity behavior, learning-based models \cite{rowe_ctrl-sim_2024, philion_trajeglish_2024, lu_scenecontrol_2024} trained on real-world data have been proposed. Current research focuses on the application of \acp{LLM} \cite{tan_promptable_2024, liu_controllable_2024, tian_enhancing_2024} to enable natural language descriptions to be mapped to low-level actions for use in traffic simulation.  
Recent reviews of traffic simulation have focused on developing frameworks for training and assessing traffic entities or evaluating various machine learning techniques, including analyses of their respective advantages and disadvantages \cite{montali_waymo_2024, chen_data-driven_2024}.
However, these reviews do not analyze the entity behavior within traffic simulations or address the mechanisms of scenario behavior control, particularly in terms of managing reactive or adversarial traffic entities. 
Furthermore, they do not consider the potential impact of behavioral control mechanisms on the training and evaluation of \ac{AV} algorithms. 
In contrast, we present a comprehensive overview of existing approaches for modeling traffic entities with adversarial and reactive behaviors.
Our contributions can be summarized as follows:
1) a classification of traffic entity behavior and classes of scenario behavior control, 
2) an analysis of open-source traffic simulators based on the traffic entity behavior classes they support, 3) a review of current research on traffic entities, which are analyzed in terms of their behavior, and 4) a summary of challenges and open questions.

\begin{figure*}[ht!]
  \vspace{3mm}
  \centering
  \includegraphics[width=0.9\textwidth]{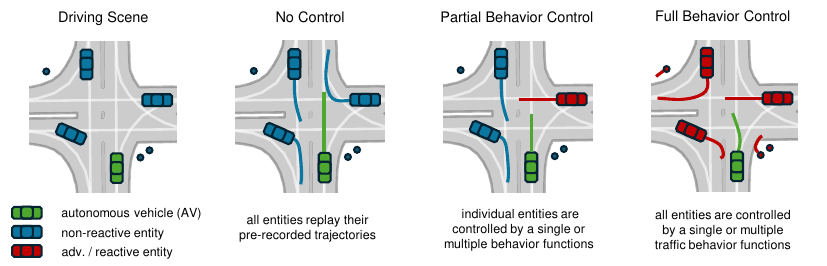}
  \caption{Proposed classification. In the absence of control over the driving scene, all entities replay their logged trajectories, while the \ac{AV} tries to maintain a safe state. In a partial controllability setting, individual entities are selected for control by a behavior function, while all other entities are set to replay. Finally, under full controllability, all entities except the \ac{AV} are controlled by a single or multiple behavior functions.}
  \label{fig:control_taxnomy}
  \vspace{-3mm}
\end{figure*}

\section{Categorization of Behavior}

According to \cite{ulbrich_defining_2015}, 
a driving scene is defined as a snapshot of the state of the environment, encompassing dynamic entities as well as the surrounding scenery, including the lane network, static entities and environmental conditions. 
A scenario is considered a temporal development of a sequence of scenes, where these scenes are linked by actions of dynamic entities (e.g. vehicles or pedestrians) based on their goals and values. 
In this context, we are particularly interested in the action selection mechanisms and the underlying value systems.  
This interest is directed toward defining different classes of behavior to better understand and categorize dynamic entities. 
Furthermore, we introduce the concept of scenario behavior control, defined by the number of entities within a scenario that possess their own decision-making systems. 
This ranges from no control, where only the \ac{AV} has its own decision-making system, to full control, where every entity is equipped with its own decision-making system and can react to surrounding entities. 

\subsection{Dynamic Entity Behavior}
\label{sec:dynamic_entity_behavior}
To classify the behavior of dynamic entities and their respective action selection mechanisms, we focus on the implementation used to simulate behavior within the traffic environment. Formally, this can be described by the following:
\begin{equation}
    a_t = f(o_t),
\end{equation}
where $f$ is a generic behavior function that maps the entity's observations $o_t$ to actions $a_t$. 
Observations represent perceptions of the traffic environment, which may include the positions of surrounding entities and map information. In contrast, actions define the motion control, such as the selected acceleration and steering.
The goal of designing a behavior function is to model an entity's dynamics and interactions with other traffic participants either realistically or in a predetermined manner.
Building on the entity definitions proposed in \cite{gulino_waymax_2024} and \cite{chen_data-driven_2024}, we distinguish the following categories of behavior for dynamic entities.


\subsubsection{Replay Behavior} 
Dynamic entities with replay behavior follow pre-recorded actions and trajectories obtained from either real-world or simulated datasets. 
These entities are non-reactive, meaning they do not adjust their behavior in response to surrounding traffic participants.
The replay behavior can be described as: 
\begin{equation}
    a_t = f_{\text{replay}}(T_t),
\end{equation}
where $T_t$ represents the trajectory demonstration, and $f_{\text{replay}}$ is the mapping function that converts these demonstrations into usable simulation outputs.
Note that observations are not considered in this representation.
Dynamic entities with replay behavior tend to overestimate the aggressiveness of real actors, as they strictly follow their planned route in all situations without deviation.

\subsubsection{Rule-based Behavior} 
Dynamic entities with rule-based behavior use stochastic functions, simple heuristics, or analytical models to determine their actions \cite{treiber_congested_2000, kreutz_analysis_2021, sugiyama_optimal_1999, kesting_general_2007}, which are defined by:
\begin{equation}
a_t = f_{\text{rule}}(o_t; \theta).
\end{equation}
The behavior function $f_{\text{rule}}$ takes the observation $o_t$ as input, along with a parameter vector $\theta$ that can be used to modify the underlying function.
These approaches can generate feasible trajectories without the need for large datasets, offering strong interpretability, which enhances the credibility of the simulation and ensures compliance with legal and ethical requirements. 
However, rule-based behavior typically follows fixed routes, offering limited reactivity and tends to be overly accommodating, often stopping to avoid collisions.

\subsubsection{Learning-based Behavior} 
These dynamic entities utilize advanced algorithms \cite{huang_cadre_2024} or deep neural networks \cite{philion_trajeglish_2024} to learn a behavior function $f_{\text{learn}, \theta}$ that better captures traffic diversity: 
\begin{equation}
    a_t = f_{\text{learn}, \theta}(o_t),
\end{equation} 
where $\theta$ defines the parameters of the trained neural network. 
These entities generate behavior that is flexible and adaptable to different scenarios. 
Furthermore, learning-based training methods can be employed to account for forced interactions between agents or to induce adversarial behavior in the scenario.
However, large amounts of training data and complex neural network architectures may be required, which can be computationally expensive and time-consuming. 
Additionally, these entities often lack transparency, making it difficult to understand how actions are chosen and behaviors generated. 
Finally, defining an appropriate objective function in mathematical form for training the neural network requires domain knowledge. 

\subsubsection{Language-based Behavior} 
Entities with language-based behavior leverage \acp{LLM}, such as GPT-4 \cite{achiam_gpt-4_2023}, to translate behavior described in natural language into low-level actions. This mapping can be expressed as:
\begin{equation}
    a_t = f_{\text{language}}(o_t;\text{LLM}),\end{equation}
where $f_{\text{language}}$ is a function that connects the observation $o_t$ and the \ac{LLM}. The resulting output can be either direct physical actions in the traffic simulation, guided by the desired entity behavior specified in the \ac{LLM}'s system prompt \cite{gao_laser_2024}, or a meta-action \cite{liu_controllable_2024}, which is further interpreted by downstream functions or networks. While the use of \acp{LLM} eliminates the challenge of defining an objective function, as a language description can be used directly, they are prone to hallucination. Moreover, using \acp{LLM} to generate low-level actions often leads to runtime issues in the simulation due to long inference times.

\subsection{Scenario Behavior Control} 
\label{sec:driving_scene_classes}
Scenario behavior control is classified based on the number of entities with decision-making systems, rather than those merely following pre-recorded trajectories. We propose the following classification, as illustrated in Fig.~\ref{fig:control_taxnomy}.

\subsubsection{No Control}
In a scenario with non-controllable behavior, all entities, except the \ac{AV}, exhibit replay behavior, strictly following their pre-recorded trajectories and actions.
The majority of simulation environments \cite{dosovitskiy_carla_2017, althoff_commonroad_2017, lopez_microscopic_2018, bernhard_bark_2020, kothari_drivergym_2021, li_metadrive_2022, vinitsky_nocturne_2022, amini_vista_2022, scott_scenario_2023, gulino_waymax_2024} that support expert or real-world data utilize replay-behavior models to simulate recorded scenarios or to test and train \ac{AV} control and planning algorithms. However, if the \ac{AV}'s actions differ from those in the recorded log, entities with replay behavior will not react accordingly, making the simulation unrealistic and ineffective for validation.

\subsubsection{Partial Behavior Control}
In a scenario with partially controllable behavior, individual entities are equipped with decision-making systems, while all other entities are set to replay \cite{philion_trajeglish_2024}.
This control schema is of particular interest when testing the performance of the \ac{AV} policy in specific contexts to ensure intended functionality, as required, for example, by the Safety of the Intended Functionality (SOTIF) \cite{international_organization_for_standardization_road_2022}. 
Additionally, UN Regulation No. 157 for active lane-keeping systems \cite{u_n_e_c_for_europe_unece_proposal_2022} defines scenarios where a single vehicle must create a forced interaction with the AV. In this setting, the repeatability is balanced against the necessary degree of freedom needed to create a variety of scenarios required for safety validation.
Partial behavior control can be divided into two distinct categories: \textit{two-objects} and \textit{mixed-traffic}. 
In a two-objects setting, scenarios consist of only a single additional entity alongside the AV. 
In contrast, in mixed-traffic settings, the scenarios are supplemented with background entities that exhibit replay behavior. 

\subsubsection{Full Behavior Control} 
In a scenario with full behavior control, all entities have their own decision making system and are able to respond to changes in the behavior of surrounding entities. 
The full controllable setting can be further classified: \textit{agent-centric} and \textit{scene-centric} \cite{xu_bits_2023}. 
In a scene-centric setting, joint predictions are generated for all (or a subset of) entities' actions within a given scene, focusing on realistic interaction behavior. 
In contrast, in agent-centric settings, each dynamic object makes decisions in a decentralized manner, without explicit coordination, using either replicas of the same behavior function or different types of behavior functions. 
The agent-centric setup is particularly well-suited for practical simulation use cases, where entities with different behavior types interact. 

\section{Traffic Simulation} \label{sec:sim_and_data}

Traffic simulations are essential components of environmental representation, as they enable the deployment of diverse traffic entities and the evaluation of  \ac{AV} algorithms. 
These simulations can be categorized into three levels \cite{xu_bits_2023, kotusevski_review_2009, ejercito_traffic_2017, schutt_taxonomy_2022}: \textit{macroscopic}, which analyzes large-scale traffic flow without detailing the states of individual entities; \textit{microscopic}, which focuses on individual traffic participants and simulates their interactions; and \textit{nanoscopic}, which models the internal decision-making processes of individual entities by subdividing behavior into subcomponents.  
Given this work's focus on individual entities and interactions, the emphasis is placed on microscopic traffic simulation. 


\begin{table}[b!]
    \scriptsize
    \setlength{\tabcolsep}{3pt}
    \centering
    \caption{Comparison of traffic simulators (chronological order)}
    \begin{tabular}{lccccccc}
        \toprule
        Simulator & \makecell{Multi- \\ entity } & \makecell{Sensor \\ Sim } & \makecell{Expert \\ Data} & \makecell{Sim- \\ entity} & \makecell{Real \\ Data} & \makecell{Traffic \\ behavior} \\
        \midrule 
        TORCS \cite{wymann_torcs_2000} & & $\surd$ & & $\surd$ & &Rule\\
        CarRacing \cite{brockman_openai_2016} & & & & & &  - \\
        CARLA \cite{dosovitskiy_carla_2017}&&$\surd$&$\surd$& $\surd$ & $\surd$ &Replay\\
        CommonRoad \cite{althoff_commonroad_2017} & & & & $\surd$ & $\surd$ &Replay\\
        Duckietown \cite{paull_duckietown_2017} & & $\surd$ & & & $\surd$ &-\\
        AirSim \cite{shah_airsim_2018} & & $\surd$ & & & $\surd$ &-\\
        HighwayEnv \cite{leurent_environment_2018} & & & & & &Rule\\ 
        Sim4CV \cite{muller_sim4cv_2018} & & $\surd$ & & & &-\\
        SUMO \cite{lopez_microscopic_2018} & & & & & $\surd$ &Replay,Rule\\
        BARK \cite{bernhard_bark_2020}&$\surd$&&$\surd$&$\surd$&$\surd$ &Replay,Rule\\
        DeepDrive-Zero \cite{quiter_deepdrive_2020} & $\surd$ & & & $\surd$ & &Learn\\
        MACAD \cite{palanisamy_multi-agent_2020}&$\surd$&$\surd$&&$\surd$&&Learn\\ 
        SUMMIT \cite{cai_summit_2020}& $\surd$ & $\surd$ & & $\surd$ & $\surd$ & Rule\\
        DriverGym \cite{kothari_drivergym_2021}&&&$\surd$&$\surd$&$\surd$&Replay,Learn\\
        L2R \cite{herman_learn--race_2021} & & $\surd$ & & & & - \\
        MADRaS \cite{santara_madras_2021}&$\surd$&$\surd$&&$\surd$&&Rule\\
        NuPlan \cite{caesar_nuplan_2021}&&$\surd$&$\surd$&$\surd$&$\surd$&Rule\\
        SMARTS \cite{zhou_smarts_2021} & $\surd$ &  & & & & Rule\\
        InterSim \cite{sun_intersim_2022}&$\surd$&&$\surd$&$\surd$&$\surd$&Learn \\
        Metadrive \cite{li_metadrive_2022} & $\surd$ & $\surd$ & $\surd$ & $\surd$ & $\surd$ & \makecell{Replay,Rule,\\Learn}\\
        Nocturne \cite{vinitsky_nocturne_2022}&$\surd$&&$\surd$&$\surd$&$\surd$&Replay\\
        VISTA \cite{amini_vista_2022}&$\surd$&$\surd$&$\surd$&&&Replay\\
        ScenarioGym \cite{scott_scenario_2023}&$\surd$&&&$\surd$&$\surd$&Replay\\
        tbsim \cite{xu_bits_2023}&$\surd$&&$\surd$&$\surd$&$\surd$&Learn\\
        TorchDriveSim \cite{lavington_torchdriveenv_2024}&$\surd$&&&$\surd$&&Learn\\
        Waymax \cite{gulino_waymax_2024}&$\surd$&&$\surd$&$\surd$&$\surd$ & \makecell{Replay,Rule,\\Learn}\\
        LimSim++ \cite{fu_limsim_2024} &$\surd$&$\surd$&&&$\surd$& \makecell{Language}\\
        \bottomrule
    \end{tabular}
    \label{tab:simulators}
\end{table}

\subsection{Microscopic Traffic Simulators}

Simulators such as CarRacing\cite{brockman_openai_2016}, Duckietown\cite{paull_duckietown_2017}, AirSim\cite{shah_airsim_2018}, Sim4CV\cite{muller_sim4cv_2018}, L2R\cite{herman_learn--race_2021} are designed to facilitate the training of basic \ac{AV} driving policies capable of controlling vehicles without the influence of surrounding entities. 
To enhance the fidelity and develop driving policies that can genuinely interact with their environment, it is essential to integrate surrounding traffic. 
Platforms like \cite{dosovitskiy_carla_2017,vinitsky_nocturne_2022, amini_vista_2022, scott_scenario_2023, althoff_commonroad_2017} support the use of dynamic entities with replay-behavior. 
The design of reactive agents is introduced through the use of rule-based behavior functions that can deviate from prerecorded plans and make local decisions, as implemented in \cite{wymann_torcs_2000,leurent_environment_2018,santara_madras_2021,caesar_nuplan_2021, zhou_smarts_2021, cai_summit_2020}. 
Learning-based entities, as used in platforms such as \cite{quiter_deepdrive_2020, palanisamy_multi-agent_2020, sun_intersim_2022, xu_bits_2023, lavington_torchdriveenv_2024}, and language-based entities \cite{fu_limsim_2024}, enable the generation of behaviors that are not constrained by predefined rules.  
To this end, many simulators have evolved into or been adapted as more general platforms that support multiple techniques. 
For example, \cite{lopez_microscopic_2018, bernhard_bark_2020} support both replay and rule-based behavior, \cite{kothari_drivergym_2021} supports replay and learning-based behavior, while \cite{li_metadrive_2022, gulino_waymax_2024} support all categories. 
A detailed comparison of the aforementioned microscopic traffic environments is presented in Table \ref{tab:simulators}. 
Here, multi-entity refers to the capability of simulating multiple entities, which enables the control of all objects within the simulator.
Sensor sim indicates the presence of implemented sensor models (e.g., camera, lidar, and radar).
Expert data pertains to human demonstrations or rollout trajectories collected using an expert policy.
Sim-entities refer to entity behavior functions used for simulating dynamic objects.
Real data denotes the ability to import real-world driving data.
Finally, traffic behavior specifies the types of dynamic entity behavior available defined in Sec.\ref{sec:dynamic_entity_behavior}.

\subsection{Traffic Datasets}

Traffic datasets provide map and trajectory data for the initialization of traffic simulators.  
The NGSIM dataset \cite{alexiadis_next_2004} is excluded from consideration as its quality is no longer considered to represent the current state of the art \cite{coifman_critical_2017}. 
The considered datasets are divided into two categories: prediction and planning.
The primary distinction between these categories is that \textit{prediction} datasets lack a baseline navigation route to indicate the high-level goals of the scenario's entities, whereas \textit{planning} datasets include a variety of scenarios and goals.
\cite{althoff_commonroad_2017} and \cite{caesar_nuplan_2021} are the only datasets in the planning category. Datasets that provide trajectories for prediction tasks are \cite{krajewski_highd_2018, chang_argoverse_2019, wilson_argoverse_2023, zhan_interaction_2019, bock_ind_2020, breuer_opendd_2020, caesar_nuscenes_2020, sun_scalability_2020, houston_one_2021, malinin_shifts_2021, gressenbuch_mona_2022}. 
A detailed comparison of the aforementioned datasets is presented in Table~\ref{tab:datasets}. 
Here, record time refers to the total duration of recorded trajectory data. 
Scenes denote the number of distinct locations captured in the dataset. 
Sampling frequency indicates how frequently trajectory positions are recorded. 
Sensor data represents the availability of sensor data and  
type refers to the category of the dataset, which can be either for prediction (Pred) or planning (Plan).

\begin{table}[!t]
    \vspace{3mm}
    \footnotesize
    \setlength{\tabcolsep}{4pt}
    \centering
    \caption{Comparison of scenario datasets  (chronological order)} 
    \begin{tabular}{lccccc}
        \toprule
        Dataset & \makecell{Record\\time} & \makecell{Scenes} & \makecell{Sampling \\ frequency} & \makecell{Sensor \\ data} & Type \\
        \midrule 
        CommonRoad \cite{althoff_commonroad_2017} & - & 39799 & 10 Hz & & Plan \\
        highD \cite{krajewski_highd_2018} & 16.5 h & 6 & 25 Hz & & Pred\\
        Argoverse 1 \cite{chang_argoverse_2019} & 320 h & 323,557 & 10 Hz & & Pred\\
        Interaction \cite{zhan_interaction_2019} & 10 h & 11 & 10 Hz & & Pred\\
        inD \cite{bock_ind_2020} & 10 h & 4 & 10 Hz & & Pred\\
        openDD \cite{breuer_opendd_2020}&  62 h & 7 & 10 Hz & & Pred\\
        nuPredict \cite{caesar_nuscenes_2020}& 5.5 h & 850 & 2 Hz & $\surd$ & Pred\\
        Waymo \cite{sun_scalability_2020} & 570 h & 103,354 & 10 Hz & &  Pred\\
        Lyft Level 5 \cite{houston_one_2021} & 1118 h & 170,000 & 10 Hz & & Pred\\
        Shifts \cite{malinin_shifts_2021} & 1667 h & - & 5 Hz & & Pred \\
        MONA \cite{gressenbuch_mona_2022} & 130 h & 3 & 5 Hz & & Pred \\ 
        Argoverse 2 \cite{wilson_argoverse_2023} & 763 h & 250,000 & 10 Hz& & Pred\\
        nuPlan \cite{caesar_nuplan_2021, karnchanachari_towards_2024} & 1282 h & - & 10 Hz & $\surd$ & Plan\\
        \bottomrule
    \end{tabular}
    \label{tab:datasets}
    \vspace{-3mm}
\end{table}

\section{Behavior-Realistic Traffic Entities} 

Adversarial and reactive traffic entities are designed to respond to or influence their surroundings based on specific objectives, with the aim of creating a realistic representation of object dynamics and interactions with other traffic participants in a traffic simulation.
Reactive traffic refers to the creation of realistic, dynamic entities that respond to the \ac{AV} in a way that closely emulates human behavior, with the goal of enabling closed-loop traffic simulations. 
In contrast, adversarial traffic focuses on simulating adversarial entity behaviors that induce realistic yet safety-critical interactions with the \ac{AV}. 
In the following, we analyze the different classes of dynamic entity behavior (Sec.\ref{sec:dynamic_entity_behavior}) with respect to reactive traffic, adversarial traffic, and a combination of both traffic modeling techniques. Additionally, we examine the defined scenario behavior control classes (Sec.\ref{sec:driving_scene_classes}).

\subsection{Rule-based Behavior Generation}
\label{sec:rule_based_agents}
Dynamic entities with rule-based behavior use analytic and heuristic models to generate reactive behaviors in response to diverging \ac{AV} plans and surrounding traffic. 
However, they are constrained to predefined trajectories and cannot adopt adversarial behavior. 
Examples include the \ac{IDM} \cite{treiber_congested_2000}, generalized \ac{IDM} \cite{kreutz_analysis_2021}, and the \ac{OVM} \cite{sugiyama_optimal_1999}, which update a vehicle’s acceleration to avoid collisions by considering the proximity and relative velocity of preceding vehicles. 
The \ac{MOBIL} model \cite{kesting_general_2007} enhances lane-change efficiency by minimizing overall deceleration or braking. 
For an overview, see \cite{brockfeld_toward_2003}.

\subsection{Learning-Based Behavior Generation}
\label{sec:learning_based_agents}
Learning-based behavior functions can generate scenarios with realistic interactions. Exemplary approaches are discussed below. A detailed comparison of these approaches regarding scenario behavior control, traffic modeling techniques, and applied learning methods is presented in Tab.\ref{tab:agents} and Tab.\ref{tab:methods}

\subsubsection{Reactive Traffic} 
An intuitive approach to reactive traffic is training neural networks to mimic the behavior of demonstrators, such as those in the datasets mentioned in Section \ref{sec:sim_and_data}, using \ac{IL}. \ac{IL}-based entities learn from demonstrations to perform tasks similar to expert human drivers. For example, \cite{zheng_learning_2020, guo_long-term_2023} address traffic simulation learning with \ac{GAIL}, while \cite{zhu_rita_2022} extends this to capture individual driving styles, and \cite{yan_learning_2023} focuses on statistical realism.
Model-based \ac{GAIL} 
is used in \cite{bhattacharyya_multi-agent_2018,bhattacharyya_modeling_2022,igl_symphony_2022} to enhance behavioral realism and diversity. \cite{xu_bits_2023} combines high-level intent inference with low-level goal-conditioned control in a bi-level \ac{IL} framework. \cite{bergamini_simnet_2021} uses \ac{BC} for modeling behavior policies.
\Ac{RL} approaches leverage human demonstrations to improve traffic behavior models. Human feedback refines \ac{IL}-based models, as seen in \cite{xu_bits_2023} and \cite{cao_reinforcement_2024}, or serves as a regularization term to guide the learning process \cite{cornelisse_human-compatible_2024}. \cite{shiroshita_behaviorally_2020} employs multiobjective \ac{RL} to balance diversity and driving skills, while \cite{zhang_trajgen_2022} focuses on optimizing collision-avoidance trajectories. Additionally, \cite{sackmann_modeling_2022} uses inverse \ac{RL} to learn driving policies from real-world data.
\Acp{DGM} use human demonstrations to learn probabilistic models, enabling new sample generation. \cite{suo_trafficsim_2021} uses a \ac{VAE} to learn a latent representation of driving data to generate realistic scenarios. Similarly, \cite{zhang_trafficbots_2023} employs a conditional \ac{VAE}, while \cite{tang_exploring_2021} combines a \ac{VAE} with sparse graph attention message-passing.
\Acp{DAR} generate explicit density models; for example, \cite{feng_trafficgen_2023} uses a generative \ac{DAR} to synthesize realistic traffic scenarios, and \cite{philion_trajeglish_2024} presents a \ac{DAR} for dynamic driving scenario generation. \cite{jiang_motiondiffuser_2023} employs a diffusion model for joint multi-agent motion prediction and introduces a constrained sampling framework for controlled trajectory generation, while \cite{zhong_guided_2023} uses a conditional diffusion model for controllable, realistic multi-agent traffic.
Finally, \cite{sun_intersim_2022} and \cite{chang_analyzing_2022} apply deep learning techniques to simulate realistic and reactive entity trajectories.

\setlength{\tabcolsep}{4pt} 
\begin{table}
\vspace{3mm}
  \scriptsize
    \caption{Classification of Learning-based Behavior}
  \label{tab:agents}
  \centering
  \begin{tabular}{c|ccc}
    \toprule
    & \makecell{Reactive \\ Traffic} & \makecell{Adversarial \\ Traffic} & \makecell{Adversarial and \\ Reactive Traffic} \\
    \midrule
    \multirow{4}{*}{\makecell{Partial \\ Behavior \\ Control}}
    &
    \makecell{
    \textit{Two-Objects}\\
    \cite{philion_trajeglish_2024},\cite{zhu_rita_2022},\cite{igl_symphony_2022},\\ 
    \cite{cornelisse_human-compatible_2024},\cite{zhang_trajgen_2022},\cite{zhang_trafficbots_2023}\\ \\
    }
    & 
    \makecell{
    \textit{Two-Objects}\\
    \cite{stoler_seal_2024}, \cite{koren_adaptive_2018},\cite{chen_adversarial_2021},\\ \cite{mavrogiannis_b-gap_2022},\cite{feng_intelligent_2021},\cite{wachi_failure-scenario_2019},\\ \cite{xu_diffscene_2023},\\
    }
    & 
    \makecell{\textit{Two-Objects}\\
    \cite{suo_mixsim_2023} \\ \\ \\}
    \\
    \\
    &
    \makecell{\textit{Mixed-Traffic}\\ \\ \\}
    &
    \makecell{\textit{Mixed-Traffic}\\ 
    \cite{rowe_ctrl-sim_2024},\cite{rempe_generating_2022},
    \cite{ransiek_goose_2024},
    \\
    \cite{huang_cadre_2024}, \cite{zhang_cat_2023}, \cite{cao_advdo_2022}, 
    \\
    }
    &
    \makecell{\textit{Mixed-Traffic}\\ \\ \\}
    \\
    \midrule
    \multirow{6}{*}{\makecell{Full \\ Behavior \\ Control}}
    & 
    \makecell{
    \textit{Agent-centric} \\
    \cite{xu_bits_2023},\cite{sun_intersim_2022},\cite{zheng_learning_2020},\\
    \cite{cornelisse_human-compatible_2024},\cite{shiroshita_behaviorally_2020},\cite{sackmann_modeling_2022},\\
    \cite{zhong_guided_2023},\cite{chang_analyzing_2022}\\
    \\
    }
    &
    \makecell{
    \textit{Agent-centric} \\
    \cite{niu_re_2023}, \cite{hanselmann_king_2022}\\
    \\ \\ \\
    }
    &
    \makecell{
    \textit{Agent-centric} \\
    \cite{suo_mixsim_2023},\cite{yin_diverse_2021},\cite{chang_controllable_2023},\\
    \cite{feng_dense_2023}\\
    \\ \\
    }
    \\
    & 
    \makecell{
    \textit{Scene-centric} \\
    \cite{bergamini_simnet_2021},\cite{philion_trajeglish_2024},\cite{guo_long-term_2023},\\
    \cite{yan_learning_2023},\cite{bhattacharyya_multi-agent_2018},\cite{bhattacharyya_modeling_2022},\\
    \cite{cao_reinforcement_2024},\cite{suo_trafficsim_2021},\cite{zhang_trafficbots_2023},\\
    \cite{tang_exploring_2021},\cite{feng_trafficgen_2023}
    }
    &
    \makecell{\textit{Scene-centric}\\ \\ \\ \\ \\}
    &
    \makecell{
    \textit{Scene-centric}\\
    \cite{lu_scenecontrol_2024},\cite{tan_scenegen_2021},\cite{jiao_tae_2022},\\\cite{ding_realgen_2023},\cite{zhong_language-guided_2023},\cite{huang_versatile_2024},\\
    \cite{rowe2025scenario}
    \\ \\
    }
    \\
    \bottomrule
  \end{tabular}
\vspace{-3mm}
\end{table}

\subsubsection{Adversarial Traffic} 

Safety-critical and adversarial behaviors are often underrepresented in data, making their generation in simulations particularly valuable. \cite{koren_adaptive_2018} introduces an \ac{RL}-based adaptive stress testing method for \acp{AV}. \cite{chen_adversarial_2021} investigates the impact of adversarial scenarios on lane-change model performance. \cite{niu_re_2023} explores adversarial policies derived from offline driving data and online simulation samples. In contrast, \cite{mavrogiannis_b-gap_2022} uses \ac{RL} to generate behavior-rich vehicle trajectories with varying levels of aggressiveness.
Systematic generation of adversarial examples is addressed by \cite{feng_intelligent_2021}, and \cite{wachi_failure-scenario_2019} presents a method for identifying failure scenarios using multi-agent \ac{RL}. \cite{ransiek_goose_2024} and \cite{zhang_cat_2023} apply goal-conditioned \ac{RL} to generate safety-critical scenarios and select plausible adversarial trajectories. \cite{rowe_ctrl-sim_2024} uses offline \ac{RL} to develop controllable entity behaviors for testing and evaluating \ac{AV} algorithms. \cite{stoler_seal_2024} combines a learned scoring function and a reactive adversary policy to increase criticality while maintaining realism.
By combining \Acp{DGM} with adversarial optimization, adversarial behaviors can be generated. For example, \cite{rempe_generating_2022} uses a \ac{VAE} to create challenging scenarios for stress-testing \acp{AV}, while \cite{xu_diffscene_2023} employs a diffusion-based framework for generating realistic, safety-critical scenarios. In \cite{huang_cadre_2024}, a Quality-Diversity algorithm is selected to generate controllable adversarial scenarios. \cite{cao_advdo_2022} introduces an optimization-based adversarial attack framework to generate realistic adversarial trajectories, and \cite{hanselmann_king_2022} proposes a method for generating safety-critical scenarios via backpropagation using a kinematic bicycle model. 

\setlength{\tabcolsep}{4pt} 
\begin{table}
\vspace{3mm}
  \scriptsize
    \caption{Methods for Learning-based Behavior Generation}
  \label{tab:methods}
  \centering
  \begin{tabular}{c|cccc}
    \toprule
    & 
    IL 
    & 
    RL 
    & 
    DGM 
    & 
    Others 
    \\
    \midrule
    \makecell{Partial \\ Behavior \\ Control}
    & 
    \makecell{\cite{zhu_rita_2022},\cite{igl_symphony_2022},\\ \\}
    & 
    \makecell{
    \cite{rowe_ctrl-sim_2024},\cite{cornelisse_human-compatible_2024},\cite{zhang_trajgen_2022} 
    \\ \cite{stoler_seal_2024},\cite{koren_adaptive_2018},\cite{chen_adversarial_2021},
    \\ 
    \cite{mavrogiannis_b-gap_2022},\cite{feng_intelligent_2021},\cite{wachi_failure-scenario_2019},
    \\
    \cite{ransiek_goose_2024},\cite{zhang_cat_2023}
    }
    & \makecell{\cite{philion_trajeglish_2024},\cite{zhang_trafficbots_2023},\cite{xu_diffscene_2023},
    \\ 
    \cite{suo_mixsim_2023},\cite{rempe_generating_2022},
    \\
    }
    & 
    \makecell{\cite{huang_cadre_2024},\cite{cao_advdo_2022}\\ 
    \\}
    \\
    \midrule
    \makecell{Full \\ Behavior \\ Control}
    & 
    \makecell{
    \cite{bergamini_simnet_2021},\cite{xu_bits_2023},\cite{zheng_learning_2020},
    \\
    \cite{guo_long-term_2023},\cite{yan_learning_2023},\cite{bhattacharyya_multi-agent_2018}
    \\
    \cite{bhattacharyya_modeling_2022}
    }
    & 
    \makecell{
    \cite{cao_reinforcement_2024},\cite{cornelisse_human-compatible_2024},\cite{shiroshita_behaviorally_2020}
    \\ 
    \cite{sackmann_modeling_2022},\cite{niu_re_2023},\cite{feng_dense_2023}
    \\ 
    \\}
    & 
    \makecell{\cite{philion_trajeglish_2024}\cite{lu_scenecontrol_2024},\cite{suo_trafficsim_2021},
    \\ \cite{zhang_trafficbots_2023},\cite{tang_exploring_2021},\cite{feng_trafficgen_2023}
    \\
    \cite{jiang_motiondiffuser_2023},\cite{zhong_guided_2023},\cite{suo_mixsim_2023},
    \\
    \cite{yin_diverse_2021},\cite{chang_controllable_2023},\cite{tan_scenegen_2021},
    \\
    \cite{jiao_tae_2022},\cite{ding_realgen_2023}, 
    \\
    \cite{huang_versatile_2024},\cite{rowe2025scenario}
    }
    & 
    \makecell{\cite{sun_intersim_2022},\cite{chang_analyzing_2022},
    \\ \cite{hanselmann_king_2022} \\}
    \\
    \bottomrule
  \end{tabular}
\vspace{-3mm}
\end{table}

\subsubsection{Adversarial and Reactive Traffic} 

Methods in this category control individual entities adversarially while simulating surrounding entities reactively. \Acp{DGM} generate new samples from learned distributions. For example, \cite{tan_scenegen_2021} uses a \ac{DAR} to create both realistic and unlikely traffic scenarios. \cite{jiao_tae_2022} incorporates behaviors such as aggressiveness and intention. \cite{ding_realgen_2023} combines retrieval-augmented generation with a spatial-temporal transformer to produce realistic, safety-critical scenarios. Similarly, \cite{lu_scenecontrol_2024} uses a transformer-based architecture with a lane \ac{GNN} to meet arbitrary constraints.
RouteGAN, proposed by \cite{yin_diverse_2021}, controls vehicles separately to generate diverse interactions, producing both safe and critical behaviors. \cite{chang_controllable_2023} develops a guided diffusion model to create safety-critical scenarios with one adversarial agent and multiple reactive agents. \cite{huang_versatile_2024} introduces versatile behavior diffusion for scenarios involving multiple traffic participants. \cite{rowe2025scenario} proposes a vectorized latent diffusion model for scene generation and an autoregressive transformer for simulating entity behaviors, enabling flexible, high-fidelity, and challenging scenarios.
\cite{feng_dense_2023} extends \cite{yan_learning_2023} by inducing individual adversarial entities.
Additionally, \cite{suo_mixsim_2023} uses \acp{GNN} to train a reactive goal-conditional policy that incorporates goals as routes on the road network. 

\subsection{Language-Based Behavior Generation} 

Natural language is used as a guiding mechanism to define entity behavior and generate corresponding outputs, enabling realistic, diverse, and controllable scenario generation.
\cite{zhong_language-guided_2023} proposes a language-guided, scene-level conditional diffusion model, complemented by an \ac{LLM}-based interface to simulate safety-critical traffic situations. 
Similarly, \cite{liu_controllable_2024} develops a diffusion-based traffic simulation that integrates \acp{LLM} for cost function design. 
\cite{tan_language_2023} uses an LLM encoder and transformer decoder, while \cite{xia_language-driven_2024} uses an LLM encoder and attention-based decoder to generate realistic and controllable scenarios from natural language.
\cite{wang2025enhancing} combines an \ac{LLM}, a generative transformer, and rule-based filtering to create user-oriented collision scenarios.
The transformation of natural language inputs into structured outputs for the creation and modification of diverse scenarios has been explored by \cite{li_chatsumo_2024}, including the generation of Python code, the use of Scenic programming \cite{zhang_chatscene_2024}, and the construction of structured XML files \cite{lu_multimodal_2024}. 
\ac{LLM}-based trajectory optimization has been used to mimic real-world driving behavior \cite{li_chatgpt-based_2024} and to generate closed-loop adversarial scenarios for training and testing \ac{AV} algorithms \cite{mei2025llm}. 
Similarly, \cite{tian_enhancing_2024} proposed a closed-loop \ac{RL} environment parameterized via an \ac{LLM}-driven curriculum learning approach. \cite{tan_promptable_2024} introduced a multimodal, promptable, closed-loop traffic simulation. 
A multi-stage \ac{LLM} pipeline with rule-based execution for generating different critical and non-critical scenarios was presented by \cite{gao_laser_2024}, while \cite{aasi_generating_2024} used a branching tree of textual descriptions to generate different out-of-distribution scenarios. 
Finally, \cite{wei_editable_2024} developed editable, photorealistic 3D driving scene simulations using several specialized \ac{LLM} agents in a collaborative workflow. 
A classification of the discussed works is shown in Tab.\ref{tab:language}. 
We compare them according to the proposed levels of scenario behavior control and the different traffic modeling techniques.

\setlength{\tabcolsep}{4pt} 
\begin{table}
\vspace{3mm}
  \scriptsize
    \caption{Classification of Language-based Behavior}
  \label{tab:language}
  \centering
  \begin{tabular}{c|ccc}
    \toprule
    & \makecell{Reactive \\ Traffic} & \makecell{Adversarial \\ Traffic} & \makecell{Adversarial and \\ Reactive Traffic} \\
    \midrule
    \makecell{Partial \\ Behavior \\ Control}
    &
    \makecell{}
    & 
    \makecell{\cite{zhang_chatscene_2024}, \cite{mei2025llm},\cite{wang2025enhancing},}
    & 
    \makecell{}
    \\
    \midrule
    \makecell{Full \\ Behavior \\ Control}
    & \makecell{\cite{tan_promptable_2024},\cite{liu_controllable_2024},\cite{xia_language-driven_2024},\\ \cite{lu_multimodal_2024},\cite{wei_editable_2024},\cite{li_chatgpt-based_2024}}
    &
    \makecell{\cite{aasi_generating_2024}}
    & 
    \makecell{\cite{tian_enhancing_2024}, \cite{zhong_language-guided_2023}\cite{tan_language_2023}, \\\cite{gao_laser_2024}}
    \\
    \bottomrule
  \end{tabular}
\vspace{-3mm}
\end{table}

\section{Challenges}

Traffic simulation offers an efficient tool for evaluating \ac{AV} planning algorithms in realistic environments. 
It mitigates the high costs and risks associated with real-world testing and serves as the first step in the deployment cycle of an \ac{AV}'s driving policy. 
However, a simulator can only fulfill its purpose if it accurately mimics real-world driving conditions and provides the necessary control over the driving scenario to function as a useful testing tool. 
 Based on the approaches reviewed and the identified needs, we have outlined the following key challenges.

\subsubsection{Behavior Realism} 
Behavior realism  is defined as the extent to which a simulation incorporates interactive traffic entities and their behavior patterns.
For a generated traffic scenario to be effective in validating an \ac{AV}'s planning algorithm, it must be sufficiently realistic to plausibly occur in the real world. 
This realism is typically interpreted as the similarity between the distributions of real-world and simulated scenarios.

\subsubsection{Behavior Diversity} 
Traffic simulations are considered behavior diverse if they include algorithms that cover a wide range of scenarios rather than focusing on a single failure instance.  
Such algorithms must demonstrate the ability to generalize to unseen situations. 
Generated traffic scenarios should not only be realistic, but also be sufficiently challenging and fulfill diverse requirements to support robust testing and evaluation of \acp{AV}. 
 
\subsubsection{Behavior Controllability} 
Controllability refers to the alignment of generated scenarios with user-specified factors or guidelines. 
In order to facilitate the generation of relevant scenarios, the traffic simulation and traffic behavior functions must support editing and customization of individual entities' behaviors and intentions through external inputs. 
Control over the scenario and the creation of specific conditions are essential for comprehensive \ac{AV} testing, thereby enhancing the role of traffic simulations in validating \ac{AV} performance.

\subsubsection{Behavior Closed-Loop}
The integration of generative models comes with an extremely high computational cost, often preventing these methods from being incorporated into the online testing loop of the \ac{AV}.
Traffic simulators should ensure that the behavior of the \ac{AV}'s driving policy is tested in a closed-loop setting rather than an open-loop setting.
A behavioral closed-loop facilitates the intermediate reaction of traffic entities to the \ac{AV}'s behavior, thereby enabling efficient testing and validation.

\section{Conclusion}

We present a detailed review of techniques for developing adversarial and reactive traffic entities, along with a comprehensive list of traffic simulators and data sources.
To classify current approaches to behaviorally realistic driving simulation, we introduce dynamic entity behavior classes and scenario behavior control levels. The levels of behavior control are determined by the number of reactive or adversarial entities involved. 
The reviewed approaches are organized by entity behavior and scenario controllability. 
Based on this review, we identify key challenges in traffic behavior generation that remain partially unsolved and call for future work.

\addtolength{\textheight}{-0.5cm}   

\bibliographystyle{IEEEtran}
\bibliography{IEEEabrv, indicies/lit, IEEEtranBSTCTL}

\end{document}